\documentclass{article}
\usepackage{spconf,amsmath,graphicx,hyperref,algorithmic,algorithm}
\usepackage{arydshln,subcaption}

\title{
CTC-DID: CTC-Based Arabic dialect identification for streaming applications
}

\name{Muhammad Umar Farooq, Oscar Saz \thanks{\copyright 2026 IEEE. Personal use of this material is permitted. Permission from IEEE must be obtained for all other uses, in any current or future media, including reprinting/republishing this material for advertising or promotional purposes, creating new collective works, for resale or redistribution to servers or lists, or reuse of any copyrighted component of this work in other works.
} } 
\address{Emotech Ltd., UK}
\begin{document}
\ninept
\maketitle
%
\begin{abstract}

This paper proposes a Dialect Identification (DID) approach inspired by the Connectionist Temporal Classification (CTC) loss function as used in Automatic Speech Recognition (ASR). CTC-DID frames the dialect identification task as a limited-vocabulary ASR system, where dialect tags are treated as a sequence of labels for a given utterance. For training, the repetition of dialect tags in transcriptions is estimated either using a proposed Language-Agnostic Heuristic (LAH) approach or a pre-trained ASR model. The method is evaluated on the low-resource Arabic Dialect Identification (ADI) task, with experimental results demonstrating that an SSL-based CTC-DID model, trained on a limited dataset, outperforms both fine-tuned Whisper and ECAPA-TDNN models. Notably, CTC-DID also surpasses these models in zero-shot evaluation on the Casablanca dataset. The proposed approach is found to be more robust to shorter utterances and is shown to be easily adaptable for streaming, real-time applications, with minimal performance degradation.
\end{abstract}
\section{Introduction}
\label{sec:intro}

Dialect IDentification (DID) is a related task to Language IDentification (LID) \cite{kulkarni2023}, where the goal is to identify different dialects of the same language with nuanced phonological, lexical and grammatical differences \cite{torgbi25}. For languages with multiple dialectal variants, this poses a significant challenge for tasks such as Automatic Speech Recognition (ASR), Machine Translation (MT) or Spoken Dialogue Systems (SDS).

Dialect, language, and speaker identification are often formulated as analogous modelling problems within the speech processing domain. Prior studies \cite{Naser2020, lonergan23_sigul} have predominantly employed x-vectors \cite{synder18} and ECAPA-TDNN \cite{desplanques20} models to address these tasks. The x-vector framework extracts fixed-dimensional utterance-level embeddings by aggregating variable-length frame-level acoustic features through TDNN blocks. ECAPA-TDNN extends this approach by incorporating channel-dependent attention mechanisms and residual connections, thereby generating more robust and discriminative utterance-level representations.

More recently, the large-scale end-to-end Whisper model \cite{whisper} has incorporated language identification as an explicit training objective. In contrast to x-vectors and ECAPA-TDNN, Whisper employs a transformer-based sequence-to-sequence architecture trained on massively multilingual speech corpora. As a result, its LID performance emerges from joint optimisation across speech recognition, translation, and language identification, enabling it to leverage both acoustic and linguistic information rather than relying solely on aggregated speaker embeddings. In previous work Whisper models have successfully been fine-tuned for the DID task \cite{adi20}. However, the larger versions of Whisper are computationally very expensive to fine-tune and run, making it less suitable for low-resource scenarios, while smaller Whisper models show performance degradation.


Previous studies have investigated the application of self-supervised learning (SSL) models for language and speaker identification tasks \cite{prasad24, fan21_interspeech}. However, these approaches typically utilise SSL models either as feature extractors for downstream LID models \cite{angra25} or use pooled representations from SSL encoders to obtain utterance-level information prior to label prediction \cite{fan21_interspeech}.


Given the limitations of prior approaches, we propose an ASR-inspired training paradigm using the Connectionist Temporal Classification (CTC) function. While the framework can be extended to both speaker, language or dialect identification, we focus on the Arabic Dialect Identification (ADI) task for empirical validation. ADI has recently attracted significant attention, as Arabic is a widely spoken language across the Middle East and North Africa, characterised by substantial dialectal variation. Considerable efforts have been devoted to developing resources in this domain, including annotated datasets and benchmark systems for dialect identification.

In contrast to conventional models that produce an utterance-level prediction, we propose an SSL-based ADI system trained to produce labels as the utterance progresses. Specifically, repeated dialect tags are employed as transcriptions for speech utterances, where the number of repetitions is determined either automatically via ASR or heuristically. With this approach, we propose the following improvements over existing approaches:

\begin{itemize}
\item Fully suitable for streaming applications, as it is not reliant on full utterances to make a prediction. 
\item Able to handle multi-speaker or code-switching scenarios, as it produces multiple labels within a single utterance.
\item Improved performance in shorter utterances, as it is less reliant on longer contexts to produce label predictions.
\end{itemize}

The empirical results presented in Section \ref{sec:res} demonstrate that CTC-DID models outperform both Whisper-medium and ECAPA-TDNN models on the ADI-17 evaluation test set, despite being trained with very limited data. Additionally, CTC-DID models achieve superior performance compared to Whisper-medium in a zero-shot evaluation on the Casablanca test set. Specifically, an mHuBERT-based CTC-DID model (119M parameters) outperforms the Whisper-medium model (769M parameters) by an absolute F1-score margin of 4.66\%. Furthermore, the results indicate that CTC-DID models exhibit greater robustness to shorter utterances. A causal streaming implementation is also presented, which demonstrates that CTC-DID models can be effectively deployed in real-time streaming scenarios, providing enhanced performance for streaming applications.

\vspace{-1em}
\section{CTC-based dialect identification system}
\label{}
\vspace{-0.5em}
In this work, we propose a self-supervised learning (SSL) based approach for dialect identification (DID), drawing inspiration from automatic speech recognition (ASR) model training paradigms. Unlike traditional DID methods that treat dialect prediction as a single-label classification task per utterance, our method predicts dialect tokens continuously across the temporal dimension of the speech signal. This aligns more closely with the token-level prediction strategy used in ASR.

To train such a model, we emulate the ASR training setup where, instead of transcribing spoken words, we repeat the dialect label for each word in the utterance. The model is trained using the Connectionist Temporal Classification (CTC) loss \cite{ctc} function. Formally, given an input feature sequence $ \mathbf{X} = (\mathbf{x}_1, \dots, \mathbf{x}_T)$, and a target sequence $\mathbf{Y} = (y_1, \dots, y_L)$, where each \( y_i \) corresponds to the dialect tag, the CTC loss enables the model to learn an alignment between the repeated dialect tags and the input sequence. 

\begin{figure}
    \centering
    \includegraphics[width=0.7\linewidth]{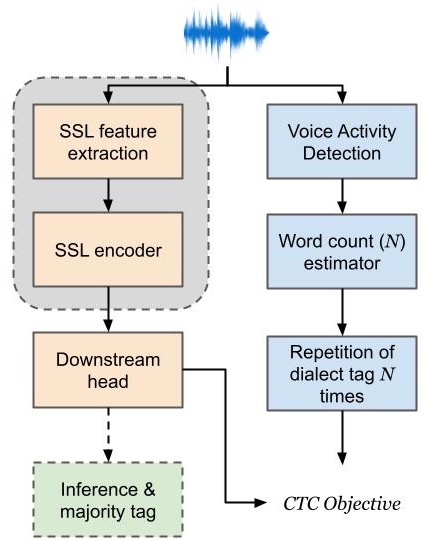}
    \caption{Flow of the proposed dialect identification approach. The components within the shaded grey box represent modules that can either be frozen or fine-tuned during training. The green box highlights the inference pipeline.}
    \vspace{-1em}
    \label{fig:flow}
\end{figure}

The flow diagram of the proposed approach is presented in Figure \ref{fig:flow}. During inference, the input utterance is decoded using the same procedure as in automatic speech recognition (ASR) models. Following the decoding step, a simple post-processing stage is applied, wherein the most frequently occurring dialect token in the decoded sequence is selected as the final dialect prediction for the utterance.

A key challenge in this formulation is determining the appropriate number of repetitions of the dialect tag, i.e., the length \( L \) of the target sequence for a given utterance. A naive yet effective approach is to estimate the number of spoken words in the utterance, and repeat the dialect token accordingly. This can be done using an off-the-shelf ASR system to count the number of words. While the accuracy of this ASR is not critical, reliance on it constrains the applicability of the method to languages supported by that ASR.

To address this limitation, we propose a lightweight and language-agnostic heuristic (LAH) for estimating the number of words. Specifically, we pass the input audio through the Silero Voice Activity Detection (VAD) \cite{SileroVAD} model to obtain the total speech duration \( d \) (in seconds). Empirical results show that assuming an average speech rate of 5 words per second (\( w = 5d \)) yields strong performance. We note, however, that this heuristic is not highly sensitive, and performance remains robust for values of \( w \geq 3d \).

Once the data is annotated with repeated dialect tags per estimated word, we train the DID model similarly to a limited-vocabulary ASR model, where the vocabulary \( \mathcal{V} \) consists of all dialect labels, along with the CTC-specific blank symbol and an optional space token.

The proposed method offers several advantages compared to traditional embedding-based approaches such as x-Vector and ECAPA-TDNN:

\begin{itemize}
    \item \textbf{Streaming Capability:} Since the model predicts dialect tokens sequentially and does not require processing the entire utterance at once, it is inherently suitable for streaming inference (Section \ref{ss:did.si}).
    
    \item \textbf{Robustness to Short Utterances:} Experimental results indicate minimal performance degradation even for very short input utterances, unlike many embedding-based methods that rely on longer temporal contexts.

    \item \textbf{Extension to Code-Switched Speech:} Due to the sequential nature of dialect token prediction, the proposed model can be naturally extended to handle code-switched speech i.e., utterances in which a speaker alternates between multiple dialects. This makes the approach well-suited for more linguistically diverse scenarios.
    
    \item \textbf{Generalisation to Speaker and Language Identification:} The same modelling framework can be readily adapted for speaker or language identification tasks. Moreover, it is expected to exhibit robustness in code-switched conditions involving multiple speakers or languages.

\end{itemize}
\vspace{-1em}
\subsection{Streaming inference}
\label{ss:did.si}
\vspace{-0.5em}


\begin{algorithm}[t]
\caption{Streaming Inference with Overlapping Chunks}
\label{alg:streaming}
\begin{algorithmic}[1]
\REQUIRE Utterance $U$, chunk duration $c$, left context $l$, sampling rate $s = 16000$
\STATE $F \leftarrow$ empty tensor
\STATE $stride \leftarrow c \cdot s$
\STATE $out\_stride \leftarrow stride / 320$
\STATE $context \leftarrow l \cdot s$
\STATE $out\_context \leftarrow context / 320$
\STATE $T \leftarrow$ total number of samples in $U$
\FOR{$t = 0$ to $T$ with step size $==stride$}
    \STATE $start \leftarrow \max(0, t - context)$
    \STATE $end \leftarrow \min(t + stride, T)$
    \STATE $chunk \leftarrow U[start : end]$
    \STATE $logits \leftarrow evaluate(chunk)$
    \STATE $frames \leftarrow logits[out\_context:]$
    \STATE $F.append(frames)$
\ENDFOR
\STATE $\hat{y} \leftarrow F.argmax(-1)$
\RETURN $\hat{y}$
\end{algorithmic}
\end{algorithm}

Traditional speaker recognition models such as x-vectors and ECAPA-TDNN rely on utterance-level representations, which inherently limit their suitability for streaming or real-time deployment scenarios. Similarly, the Whisper model processes fixed-length audio segments of 30 seconds, making it computationally expensive for streaming inference due to the overhead introduced by large input windows. While Whisper can be adapted to emulate streaming behaviour, it would need every chunk to be 30 seconds, which is computationally very expensive.

Self-supervised learning (SSL) encoders, which typically leverage Transformer-based architectures, also face similar challenges. The self-attention mechanism requires access to the entire input sequence, making it inherently non-causal and unsuitable for real-time inference without modification.


To evaluate the performance of our SSL-based dialect identification system under real-time constraints, we implement a pseudo-streaming inference approach based on overlapping audio chunks. Specifically, each utterance is segmented into fixed-length chunks of duration $c$ seconds, with an additional left-context window of $l$ seconds prepended to each chunk. This left context provides temporal continuity across chunks and enables the model to make context-aware predictions.

Given that the mHuBERT encoder downsamples the input audio by a factor of 320, an input chunk of length $c$ seconds (sampled at 16 kHz) produces approximately $50c$ output frames. For each chunk, the model outputs frame-level logits; however, only the logits corresponding to the $c$-second chunk (excluding the left-context region) are retained. These are then concatenated with the outputs of the previous chunks to form the full utterance logits. Decoding is performed over the aggregated logits. In this work, we restrict our evaluation to greedy decoding for efficiency, though beam search can be integrated for improved performance. Algorithm \ref{alg:streaming} summarises the streaming inference process.

\vspace{-1em}
\section{Experimental Setup}
\label{sec:es}
\vspace{-0.5em}
\subsection{Data}
\label{ss:es.data}
\vspace{-0.5em}
For the training of our dialect identification system, we use the ADI-17 dataset \cite{adi17} which contains 3,033 hours of 17 dialects of spoken Arabic in its train set. The test and validation sets are 2 hours per dialect. However, the scope of this work is restricted to limited resources. So, we experiment with limited amount of data i.e. 10 hours per dialect and 50 hours per dialect for training. However, the dataset is quite imbalanced and some dialects do not have 50 hours of data. In this case, for Yemeni, Moroccan and Jordan dialects, we have generated augmented data with speed perturbation and then randomly selected perturbed data to augment data to 50 hours (or 40 hours for Jordan dialect)

For development, a balanced validation set is used with 30 minutes from each dialect. For evaluation, all systems are evaluated on the full ADI-17 test set and then a zero-shot evaluation is done on the Casablanca data set \cite{casablanca} which contains around one hour per dialect.
\vspace{-1em}
\subsection{Modelling}
\label{ss:es.model}
\vspace{-0.5em}
As shown in our previous work \cite{sage}, mHubert model outperforms other SSL models such as wav2vec2.0 (also XLSR) and Hubert etc, so the mHubert model was chosen as our base SSL model.
For dialect identification, a transformer based downstream head is used. The downstream head contains 4 transformer blocks with model dimension 768, inner dimension 2048 and 8 attention heads, resulting in a 25-million-parameter head. Two alternatives are evaluated using this setup. The first one, where the transformer head is trained on top of frozen mHubert encoder weights; and, the second one, where the mHubert weights are jointly finetuned with the transformer head.



The proposed approach is compared against the ECAPA-TDNN and Whisper models. The ECAPA-TDNN model takes FBANK features as input. It consists of five layers, with the first four layers being 1D convolutional layers, each containing 1024 channels, and the final layer consisting of 3072 channels. These layers utilise varying kernel sizes and dilations to capture different temporal patterns, while 128 attention channels are incorporated to enhance feature focus. The output from the embedding model is passed through a fully connected layer with 192 neurons, followed by a classification layer. For feature normalisation, sentence-level mean and variance normalisation is applied.

Although the performance of the proposed approach is compared with a fine-tuned Whisper-base model for non-streaming configurations, the Whisper model cannot be used in streaming mode.
\vspace{-1em}
\subsection{Streaming}
\label{ss:es.streaming}
\vspace{-0.5em}

As outlined in Section \ref{ss:did.si}, the proposed approach is also evaluated for streaming speech. For experimentation, various combinations of chunk sizes and context window lengths are tested. Specifically, the streaming implementation described in Section \ref{ss:did.si} explores chunk sizes and context windows of 0.5, 1.0, 2.0, and 4.0 seconds. The results are interpolated and plotted to illustrate the performance across these different configurations.

\vspace{-1em}
\section{Results and discussion}
\label{sec:res}
\vspace{-0.5em}
\begin{table}[]
    \centering
    \caption{F1-weighted scores for the ADI-17 and Casablanca datasets with different amount of training sets. Results for the Casablanca dataset are in a zero-shot setting. The bold numbers represent the overall best performance, and the underlined values indicate the best performance within each configuration.}
    \begin{tabular}{lcc}
    \hline \hline
         & ADI-17 & Casablanca \\
    \hline
    10-hour (per dialect) training\\
    \hline
         Whisper-medium \cite{adi20} & \underline{92.88} & - \\
         ECAPA-TDNN & 28.71 & 10.18\\
         Whisper-base & 65.05 & 32.23  \\
         CTC-DID & 77.34 & 51.36 \\
         ~ + fine-tuned SSL & 86.98 & \underline{56.02}\\
    \hline
    50-hour (per dialect) training\\
    \hline
    Whisper-medium \cite{adi20}& 95.29 & -\\
    CTC-DID & 93.58 & 58.12\\
         ~ + fine-tuned SSL & \textbf{96.01} & \textbf{60.23} \\
    \hline
    Full-data training\\
    \hline
    Whisper-medium \cite{adi20} & 95.46 & 53.84\\
    Hubert \cite{casablanca} & - & 39.24 \\
    \hline
    
    \hline
    \end{tabular}
    \label{tab:results}
    \vspace{-1.5em}
\end{table}

For the initial experiments, 10 hours of data from each dialect of the ADI-17 dataset are used to either train or fine-tune the models. All models are trained for approximately 100K steps, and the results are tabulated in Table \ref{tab:results}. Model performance is evaluated on both the ADI-17 and Casablanca test sets. As demonstrated in the results, the CTC-DID approach outperforms both the ECAPA-TDNN and Whisper-base models, despite being trained with only 10 hours of data. Notably, the F1-score achieved by CTC-DID is comparable to that of the Whisper-medium model, which has a significantly larger parameter size (769M parameters). Furthermore, the evaluation on the Casablanca test set is conducted in a zero-shot setting, as no data from the Casablanca training set was used during model training. In this scenario, the CTC-DID model consistently outperforms all other systems, including Whisper-medium and HuBERT, both of which are considerably larger models. A similar CTC-DID model is trained using 50 hours from each dialect, as described in Section \ref{ss:es.data}. This model outperforms the Whisper-medium model trained on the same amount of data and even exceeds the performance of the Whisper-medium model trained on the full ADI-17 training set ($\sim$3000 hours).

For each CTC-DID model, two training approaches are considered: freezing the SSL model (CTC-DID rows in Table \ref{tab:results}) and fine-tuning the SSL model along with the downstream head (``+ fine-tuned SSL" rows in Table \ref{tab:results}). While CTC-DID models with a fine-tuned SSL encoder outperform those where the SSL encoder is used solely as a feature extractor, the latter still demonstrates competitive performance with just training the downstream head.



For the training of the CTC-DID model evaluated in Table \ref{tab:results}, the training data are prepared using the LAH approach to determine the number of word repetitions for dialect tags. To assess the performance of the data prepared through the LAH approach in comparison to data prepared using an ASR system, a model is trained using both data preparation methods. In this experiment, the model is trained with 10 hours of data without fine-tuning the mHuBERT model. The results are presented in Table \ref{tab:comapre}. As shown, the model trained on data prepared using the LAH approach demonstrates that this method does not significantly impact the results, with performance remaining comparable and effective.

\begin{table}[t]
    \centering
    \caption{Comparison of F1-weighted scores for CTC-DID models trained on data prepared using ASR-based and language-agnostic heuristic (LAH) approaches}
    \begin{tabular}{lcc}
    \hline \hline
         Data prep.&ADI-17&Casablanca\\
    \hline
         LAH & 77.34 & 51.36\\
         ASR & 79.35 & 51.84\\
     \hline
    \end{tabular}
    \label{tab:comapre}
    \vspace{-1em}
\end{table}

\vspace{-1em}
\subsection{Robustness to Short Utterances}
\begin{figure}[t]
    \centering
    \includegraphics[width=0.8\linewidth]{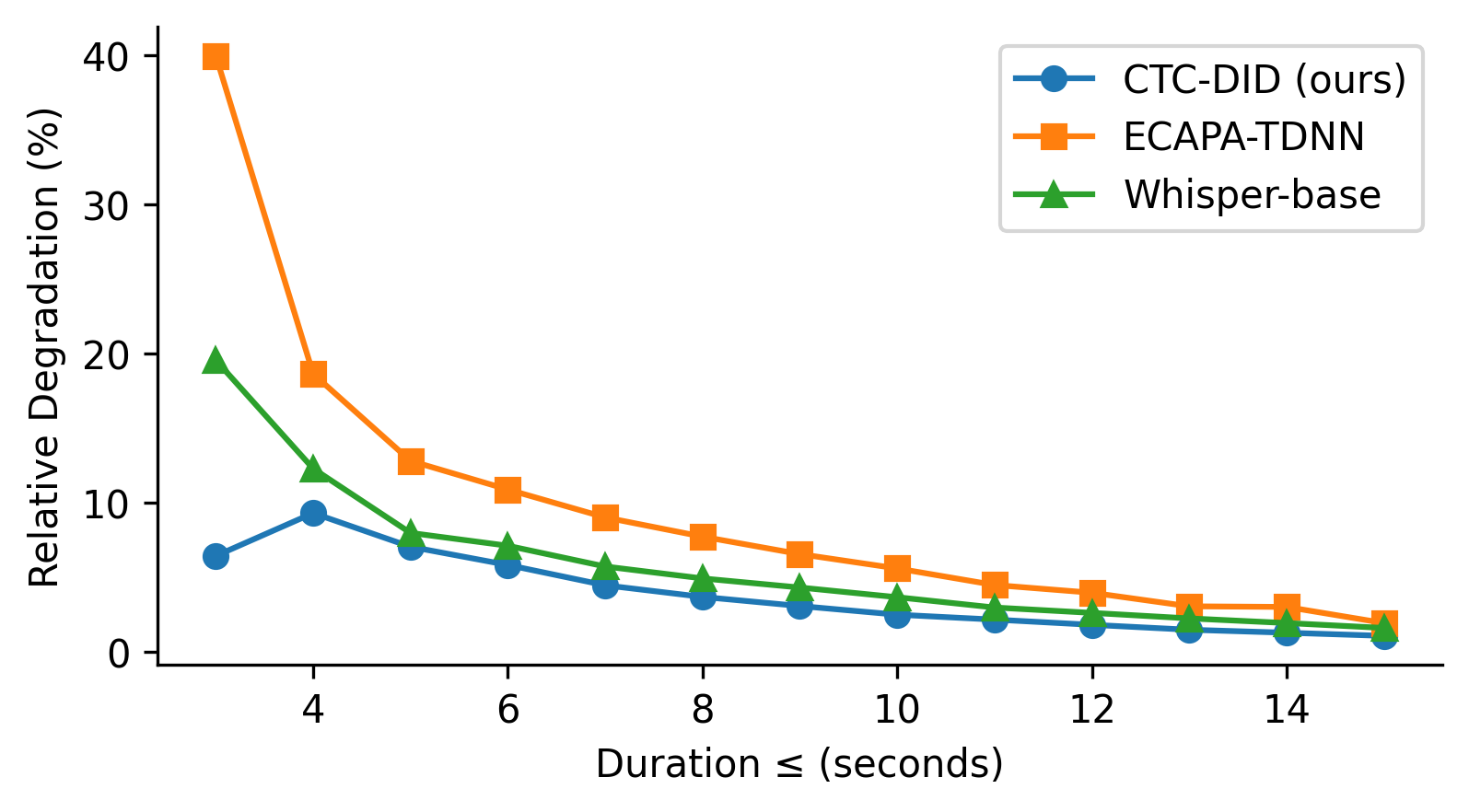}
    \caption{The relationship between relative degradation in F1 score when only the utterances of length $\leq Duration$ are evaluated through the best model.}
    \label{fig:relative}
\end{figure}

\vspace{-0.5em}
The performance of conventional utterance-level identification systems, such as those for language, speaker, or dialect identification, typically degrades for shorter utterances due to the limited availability of informative features. Given that the CTC-DID model is trained at the frame level, it is expected to exhibit greater robustness to shorter utterances. However, since the competitor models in our experiments have already been outperformed by CTC-DID, their performance on short utterances is anticipated to be suboptimal. To assess the relative robustness of CTC-DID compared to Whisper and ECAPA-TDNN, all models are evaluated on utterances with durations less than or equal to a specified threshold (in seconds), where the threshold was varied from 3 to 30 seconds. The threshold range is selected because all utterances in the ADI-17 test set are greater than 2 seconds. The relative degradation in F1-score for each threshold bin is computed for all models and is depicted in Figure \ref{fig:relative} (up to 15 seconds). As shown, CTC-DID exhibits the lowest relative degradation across all bins, indicating its superior performance robustness on shorter utterances.

\vspace{-1em}
\subsection{Streaming}
\label{ss:res.streaming}
\vspace{-0.5em}

\begin{figure}[b]
    \centering

    \begin{subfigure}[b]{0.45\linewidth}
        \centering
        \includegraphics[width=\linewidth]{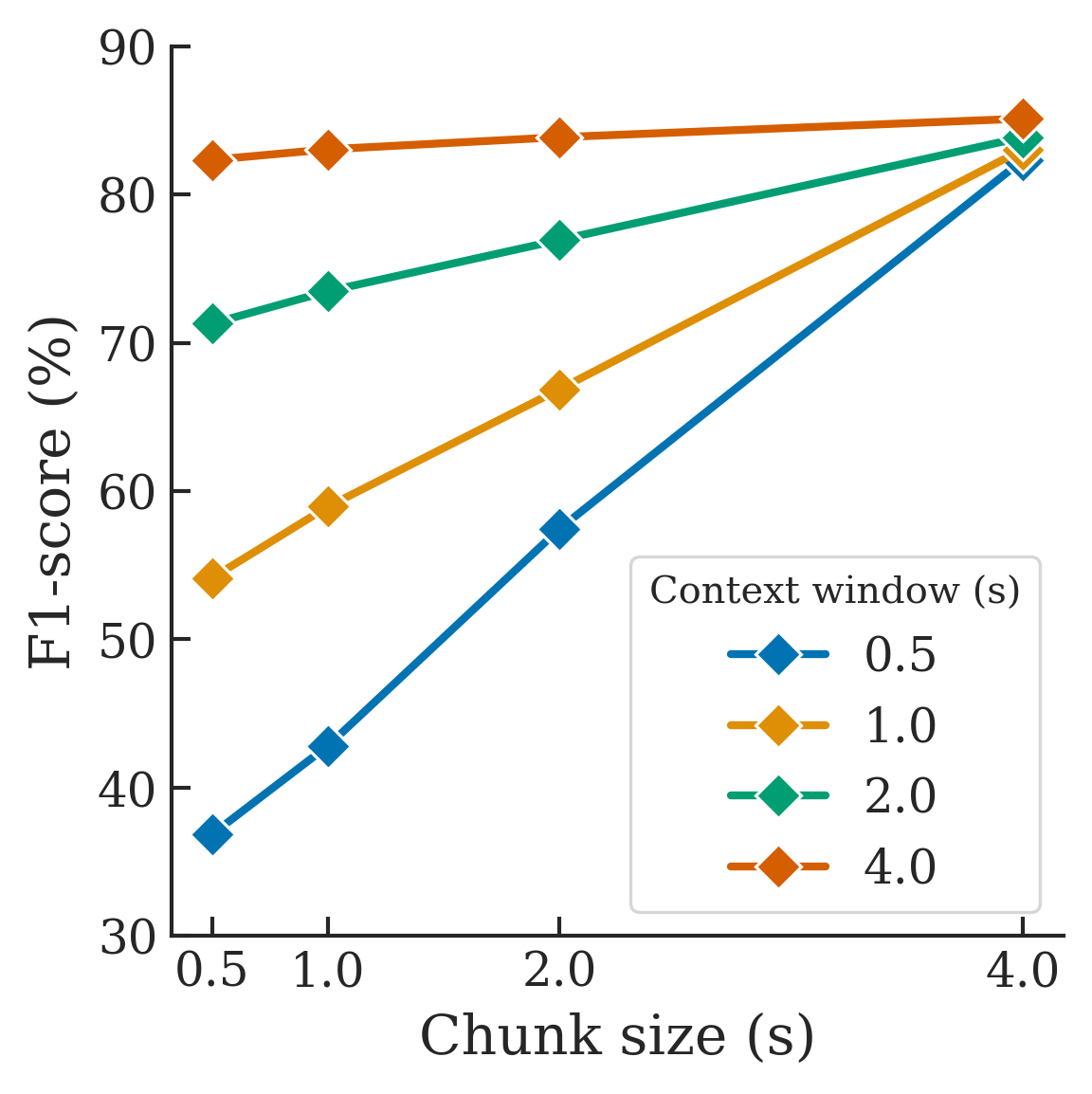}
        \caption{F1-score vs. chunk size for different past context widths}
        \label{fig:sub1}
    \end{subfigure}
    \hfill
    \begin{subfigure}[b]{0.45\linewidth}
        \centering
        \includegraphics[width=\linewidth]{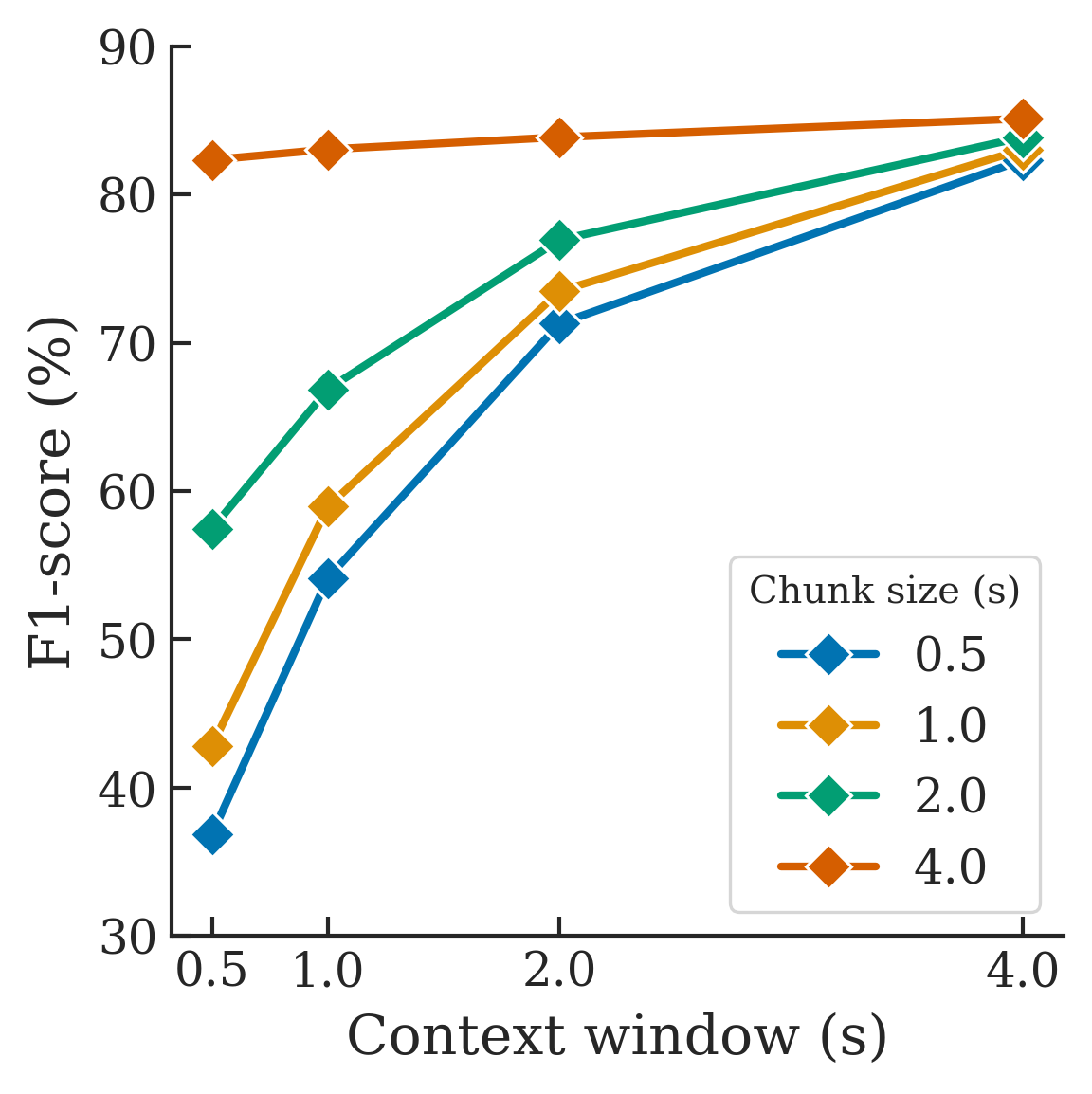}
        \caption{F1-score vs. past context widths for different chunk sizes}
        \label{fig:sub2}
    \end{subfigure}

    \caption{Effect on F1-score with different chunk sizes and context windows for streaming inference}
    \label{fig:streaming}
\end{figure}

Finally, CTC-DID models are employed for streaming dialect identification inference. Due to using the CTC function during training, the proposed model is capable of producting continuous dialect tags without requiring the full utterance. As described in Section \ref{ss:es.streaming}, a causal streaming inference mechanism is implemented to simulate real-world conditions. However, this approach inherently involves a trade-off between chunk size and performance. To investigate the impact of chunk size and the preceding (or left) context window, various combinations of chunk sizes and context windows are evaluated to determine the optimal configuration for the model. The relationship between F1-score and context window size for different chunk sizes is presented in Figure \ref{fig:sub1}. The results indicate that increasing the context size leads to an exponential improvement in F1-score for a fixed chunk size. Figure \ref{fig:sub2} illustrates the relationship between F1-score and chunk size for varying context window sizes, where the slopes of the curves appear more linear. Notably, selecting a total window size (comprising both chunk size and context window) greater than 4 seconds yields an F1-score of 82.34, which is in close proximity to the full-utterance F1-score of 86.98 (Table \ref{tab:results}).

\vspace{-1em}
\section{Conclusion}
\label{sec:conclusion}
\vspace{-0.5em}
In this work, a CTC-based dialect identification (CTC-DID) approach has been introduced, inspired by an automatic speech recognition (ASR) model training. By framing dialect identification as a limited-vocabulary ASR problem, where dialect tags are treated as words, a novel system is developed that can be trained with limited data. Our experimental results on the low-resource Arabic dialect identification task show that the SSL-based CTC-DID model outperforms established models, such as Whisper and ECAPA-TDNN, even when trained on minimal data. Furthermore, CTC-DID also demonstrates superior performance in zero-shot evaluation on the Casablanca dataset, highlighting its generalisation capability, and is more robust to shorter utterances compared to other models. The flexibility of the proposed method is further underscored by its suitability for real-time causal streaming applications, where it maintains minimal performance degradation. Although this work focuses on dialect identification, the proposed approach can be extended to other predictive tasks, such as speaker and language identification. Additionally, the method proves valuable for training systems that need to handle code-switching or multiple speakers, making it versatile for a range of speech processing applications.


\bibliographystyle{IEEEtran}
\bibliography{refs}

\end{document}